\documentclass[11pt,a4paper]{article}
\usepackage[hyperref]{acl2021}
\usepackage{times}
\usepackage{latexsym}
\usepackage{xcolor}
\usepackage{amsthm}
\usepackage{tipa}
\makeatletter
\g@addto@macro{\UrlBreaks}{\UrlOrds}
\makeatother

\aclfinalcopy

\title{gENder-IT: An Annotated English--Italian Parallel Challenge Set for Cross-Linguistic Natural Gender Phenomena}

   \author{{\bf Eva Vanmassenhove\textsuperscript{1}, Johanna Monti\textsuperscript{2}}  \\ \\
  \textsuperscript{1}Department of CSAI, Tilburg University\\
  \textsuperscript{2}UNIOR NLP Research Group, University of Naples L'Orientale \\
  {\small \tt e.o.j.vanmassenhove@tilburguniversity.edu, jmonti@unior.it}\\

  } 

\date{}

\begin{document}
\theoremstyle{definition}
\newtheorem{exmp}{Example}[section]
\maketitle
\begin{abstract}
Languages differ in terms of the absence or presence of gender features, the number of gender classes and whether and where gender features are explicitly marked. These cross-linguistic differences can lead to ambiguities that are difficult to resolve, especially for sentence-level MT systems. The identification of ambiguity and its subsequent resolution is a challenging task for which currently there aren't any specific resources or challenge sets available. In this paper, we introduce gENder-IT, an English--Italian challenge set focusing on the resolution of natural gender phenomena by providing word-level gender tags on the English source side and multiple gender alternative translations, where needed, on the Italian target side.

\end{abstract}

\section{Introduction}

Cross-linguistic differences between languages often require implicit information in the source to be made explicit on the target side. When faced with systematic structural differences between the source and target languages, human translators rely on the (broader) context (linguistic, extra-linguistic, world-knowledge) in order to infer the necessary information and adapt the target side accordingly. 

One such way in which many languages systematically differ is in terms of grammatical gender. Languages not only differ in terms of the absence or presence of specific gender features but also in the number of (linguistic) gender classes, how and where gender features are marked, and in the underlying rules by which gender is assigned~\cite{audring_2016}.\footnote{Linguistic gender classes can (and often do) correspond to what is referred to in linguistics as the natural gender of referents (i.e. `masculine' and `feminine'). However, within the field of linguistics the term `gender class' is somewhat confusing as it is often used as a synonym for noun class. There are, for instance, language with more than 3 gender classes (e.g. Kiswahili has 9) as the classes are based on different semantic distinctions. Likewise, there are languags with only two `gender classes' which correspond e.g. to an animate vs inanimate distinction.} 

In languages with grammatical gender all nouns have an (arbitrarily) assigned lexical gender.\footnote{The Dutch word for 'sun' is `zon' (female), while the French word for `sun' is `soleil' (male).} In most cases, the lexical gender of a noun is covert and can only be inferred from the morphological agreement with other words (articles, verbs, adjectives...)~\cite{Corbett1991}. However, when nouns refer to animate referents, overt gender markings corresponding to the so-called `natural gender' (biological sex) of the referent are common (e.g. the Spanish word for `nurse' is `enfermero' (male) or `enfermera' (female)).  Such forms are generated using derivational suffixes and are often derived from the `generic male'. This process is sometimes denoted as `female marking'~\cite{doleschal2000gender,laleko2018difficult}.

While language learners encounter difficulties memorizing the lexically stored gender of foreign nouns~\cite{rogers1987learners}, Machine Translation (MT) technology, given the limited (linguistic and extra-linguistic) context most MT tools leverage, struggles with the explicitation of ambiguous forms, i.e. the process of disambiguation. So far, little research has been conducted on controlling the output of MT systems in terms of features such as gender and/or number that arise due to specific cross-linguistic differences. We believe that there are two main reasons for this: (i) The research that has been conducted in this area shows that controlling specific features is a technically very challenging problem. Especially given the fact that it often requires in-depth linguistic knowledge and specialized linguistic tools, the performance of the latter often depending on how well-researched and well-resourced the languages in question are; (ii) The lack of high-quality, human-crafted challenge sets that target specific cross-linguistic phenomena. 

In this paper, we present \textit{gENder-IT}\footnote{The dataset is publicly available under a CC BY-NC-ND 3.0 through: ~\url{https://github.com/vnmssnhv/gENder-IT}.}: a word-level (human) annotated, adapted and cleaned version of a subset (English-Italian) of the MuST-SHE corpus version 1.0\footnote{Version 1.2 released more recently adds some additional information such as a `genderterms' column where pairs of corresponding gender-marked words from correct/wrong reference translations are provided}~\cite{bentivogli2020} The main contribution of our work is threefold: (i) We provide simple word-level annotations for all nouns and pronouns referring to human referents for the English sentences; (ii) While the transcripts of MuST-SHE are accompanied with gender information (male, female) of the speaker, our word-level tags can be either male or female, but also ambiguous, when the sentence itself does not provide any explicit clues with respect to the gender of the referents; (iii) We focus on the English textually gender-ambiguous sentences and provide all the correct\footnote{We ought to note that the MuST-SHE corpus provides wrong gold reference translations with swapped masculine/feminine forms of gender phenomena  - in some case, these so-called `wrong-references' correspond to the alternatives we provide.} gender-alternative translations for Italian.  

The main motivations behind our work are the following: (i) First of all, there is a need for controlled diversity within the field of MT when it comes to controlling specific features of translations, specifically when dealing with structural cross-linguistic differences ~\cite{Vanmassenhove2020}. To allow for controlled diversity, we created the first test set that allows research on identifying ambiguity and generating multiple translation variants in terms of gender; (ii) Second, recent work by Saunders et al.~\citeyearpar{saunders-etal-2020-neural} indicates that even a (very) small synthetic set of high quality sentences annotated for gender can be leveraged to improve the accuracy of translations in terms of gender specific phenomena without decreasing the overall quality. Their work was limited to annotations for one referent per synthetic sentence and focused specifically on debiasing data in terms of gender. As highlighted in Vanmassenhove et al. \citeyearpar{vanmassenhove2019getting} and Saunders~\citeyearpar{saunders-etal-2020-neural}, the effects of specific interventions need to be carefully examined on test sets that capture the complexity of a problem to its full extent. The manually annotated test set created does so by relying on `natural' (as opposed to synthetic) data that is not limited to a single human referent per sentence.   
\\
\paragraph{Bias statement}\citep{blodgett-etal-2020-language} \\In summary, this dataset is intended to encourage work on gender bias in MT, but could equally be leveraged for monolingual research on the generation of gender diverse translations (in Italian) and gender identification of referents (for English). The detailed analysis on English-Italian is intended to raise awareness on cross-linguistic differences between languages in terms of gender. NLP technologies are prone to the perpetuation (and possibly also the exacerbation \citep{vanmassenhove-etal-2021-machine}) of inappropriate stereotypes and are currently unable to recognize or warn the user about the (gender) assumptions that have been made (e.g. by translating ambiguous source sentences systematically into one specific gendered variant on the target side). Furthermore, current systems lack the ability for the user to indicate and/or control the gender of referents if needed. As such, the gender of referents in the generated MT output depends entirely on the training data which might contain (un)conscious biases that are transmitted in (written and spoken) datasets.

\section{Related Work}

Recent years, several datasets were created that focus specifically on gender-related issues observed in (sub)fields of Natural Language Processing (NLP). Targeted gender datasets (test sets or corpora) exist for subfield such as coreference resolution~\cite{rudinger2018gender, zhao-etal-2018-gender, webster2018mind} and sentiment analysis ~\cite{kiritchenko-mohammad-2018-examining}.  In this section, we will limit our discussion to datasets created specifically for mitigating and assessing gender bias in MT.

In the field of MT, \citet{mirkin-meunier-2015-personalized} used a recommender system approach to predicted user-based preferred translations based on preferences of similar users.
Rabinovich et al. \citeyearpar{rabinovich-etal-2017-personalized} worked on personalized Statistical MT. Their work centers around the preservation of gender traits by treating gender as a separate domain. For their experiments, they created a bilingual parallel corpus (English--French and English--German) annotated, among others, with the gender of the speaker.\footnote{The dataset is publicly available: \url{http://cl.haifa.ac.il/projects/pmt/}} For Neural MT, \cite{vanmassenhove2019getting, vanmassenhove2018europarl} experimented with the integration of speaker gender-tags added to the source side of the parallel corpus. Using the demographic information released by Rabinovich et al. \citeyearpar{rabinovich-etal-2017-personalized}, they compiled large datasets with gender information for 20 language pairs.\footnote{\url{https://github.com/evavnmssnhv/Europarl-Speaker-Information}} Both papers~\cite{rabinovich-etal-2017-personalized, vanmassenhove2019getting} focused specifically on gender agreement with the first person singular. As such, their corpora are limited to sentence-level gender-tags indicating the gender of the speaker.\\
Stanovsky et al.~\citeyearpar{stanovsky-etal-2019-evaluating} presented ``WinoMT'' a challenge set for the evaluation of gender bias in MT. The set is based on two existing data sets for gender bias in coreference resolution: WinoBias~\cite{zhao-etal-2018-gender} and Winogender~\cite{rudinger2018gender}. WinoBias and Winogender consist of English sentences with two human entities in the form of two gender-neutral occupations (e.g. 'teacher', 'mechanic','assistant'...) and a gendered pronoun referring to one of the two human referents. WinoMT is a concatenation of WinoBias and Winogender and contains a total of 3,888 synthetic English sentences balanced for gender. The main contribution in \citet{stanovsky-etal-2019-evaluating} is an automatic evaluation of six popular MT systems on eight language pairs.\footnote{English being the source language, and French, Italian, Russian, Ukranian, Hebrew, Armenian and German as target languages.} They provide an automatic gender bias evaluation protocol and show that the level of agreement with human annotations is above 85\% for all languages.\\
\citet{costa-jussa-etal-2020-gebiotoolkit} presents the `GeBioToolkit', a toolkit for the extraction of gender-balanced multilingual corpora with document-level gender annotations. They also introduce two versions of the `GeBioCorpus'. The first one contains 16k sentences used for evaluating the automatically extracted parallel sentences. From the evaluation, it resulted that the human annotators gave the tool on average a 87.5\% accuracy. The second version is a high-quality non-synthetic set of 2k English, Spanish and Catalan sentences post-edited by native speakers.

\citet{saunders-byrne-2020-reducing} created a small hand-crafted set of gender-balanced sentences for model adaptation. The set consists of 388 English synthetic sentences containing professions and their manually generated translations in each target language (Hebrew, German and Spanish). \citet{saunders-etal-2020-neural} explore the potential of explicit word-level gender inflection tags showing promising results. As such, gender tagging could be an effective tool for automatic translation tools where the user could specify the desired gender of the referents.

Our English-Italian parallel challenge set contains natural sentences (as opposed to synthetic) that do no follow a specific pattern\footnote{\citet{saunders-byrne-2020-reducing,zhao-etal-2018-gender} use synthetic sentence generated using templates such as ``[entity1] verb [entity2]...'' or ``The [profession] verb [pronoun] noun''.} with word-level gender inflection tags. Since naturally occurring sentences are more complex and can contain multiple entities, animate nouns and pronouns have been annotated with word-level tags that indicate the gender given the limited sentence-level context. Unlike previous work, the challenge set is not limited to specific phenomena (e.g. 1$^{st}$ or 3$^{rd}$ person singular) but covers the full range of natural gender phenomena. It is specifically designed to encourage work on controlling output in terms of gender, the identification of gender ambiguous sentences and co-reference resolution.

\section{Creation and Annotation of Dataset}
In this section, we describe the pre-processing, cleaning and the gender annotations steps.
\subsection{MuST-SHE}
The gENder-IT challenge set is based on the MuST-SHE corpus (version 1.0) comprising of naturally occurring sentences retrieved from TED Talks. We limited ourselves to the EN-IT parallel data and focused on data pertaining to what is referred to as `category 2,3 and 4' in MuST-SHE, which are defined as sentences that contain contextual hints such as
gender-exclusive words ( mom ), pronouns ( he,she ) or proper nouns ( Paul ) that inform
about the gender of the referent (category 2), sentences where both the audio signal and utterance context are needed to disambiguate the gender of referents (category 3) and sentences without contextual (audio or textual) gender information for disambiguation (category 4). 


\subsubsection{Corpus cleaning}
While MuST-SHE contains segments (one or multiple sentences), we treated every sentence independently given that most state-of-the-art MT systems work on the sentence level. Aside from splitting the segments, sentences for which the target or source part was missing were removed, spelling mistakes corrected, and missing quotations marks and punctuation were added where missing. In total, 695 sentences were annotated with word-level gender information on the English side producing a total of 314 alternative Italian translations for 114 gender-ambiguous English sentences.

\subsection{Word-level gender tags}\label{subsec:word-leveltags}
\paragraph{Annotations} Word-level gender annotations are provided for all (pro)nouns referring to a person with exception to the few nouns in English that are already gender specific.\footnote{Either due to the form: `waitress', `actress' or because of semantic features: `mother', `brother'...} In example ~\ref{ex:multient}, the (pro)nouns are tagged with their respective genders based on the textual context, except for the noun `dad'. The tags provided are $<$F$>$ or $<$M$>$ when it is clear from the sentential context that the referent should be referred to with male/female pronouns (see Ex. \ref{ex:multient}).

\begin{exmp}
\label{ex:multient2}
`So she turned and she looked at her dad, and she said, ``Dad, I $<$F$>$ know how you $<$M$>$ feel, but I $<$F$>$ don't believe in the death penalty.'''
\end{exmp}

\noindent
In all other cases, the $<$A$>$ tag is used to indicate that within the given context, no assumption can or should be made with respect to the gender of the referent. When there are multiple $<$A$>$ tags, we further distinguish between $<$A1$>$, $<$A2$>$, etc. to indicate that different entities are being referred to. This is important from a translational point of view, since it could imply that more than two translations need to be generated. For instance, in  the following sentence (Ex. \ref{ex:multient}), two nurses ($<$A1$>$ and $<$A2$>$) are mentioned refering to two different entities of which the gender, within this particular context, is unknown. In Italian, there is a male and female form for the English word \textit{nurse}: infermiera (female) and infermiere (male), which implies that there are at least four correct translation alternatives in terms of gender.


\begin{exmp}
\label{ex:multient}
``And it was there that another nurse $<$A1$>$, not the nurse $<$A2$>$ who $<$A2$>$ was looking after Mrs. Drucker $<$F$>$ before, but another nurse $<$A1$>$, said three words to me $<$A3$>$ that are the three words that most emergency physicians $<$A4$>$ I $<$A3$>$ know dread.''
\end{exmp}

Usually, annotating (pro)nouns suffices to indicate the contextual natural gender of referents, however in some cases, nounless adjectives can appear that refer to a human referent. Therefore, adjectives functioning as nouns (e.g. `the rich'...) and/or adjectives used in a (conversational) constructions without a (pro)noun (e.g. `And sporty $<$A$>$!) were tagged as well.

\paragraph{Proper names}
Many of the gender clues within the textual context referred to in the MuST-SHE corpus depend on the names of referents mentioned within the context. We opted for a slightly different approach in terms of proper names given the variety of naming conventions that exist in different cultures. Furthermore, a person's pronoun preferences might not necessarily match with the gender we traditionally or prototypically associated with a name. As such, proper names by themselves are not considered a gender clue (see Ex. ~\ref{ex:propname1}).
\begin{exmp}
\label{ex:propname1}
``Vera $<$A$>$ was dead.'' 
\end{exmp}
\noindent
We make an exception for cases where the full name of a person is given and this person can be considered a `public figure' for whom the pronouns can be retrieved, see Ex.~\ref{ex:propname2}.\footnote{In practice and for consistency, we verified whenever a full name was given whether the referent has a Wikipedia page on which they are being referred to with specific pronouns.}
\begin{exmp}
\label{ex:propname2}
`The German physicist $<$M$>$ Werner Heisenberg $<$M$>$ said, ``When I $<$M$>$ meet...  ''.'
\end{exmp}

\noindent
In total, 950 word-level tags are provided of which 138 are $<$F$>$ (15\%), 190 $<$M$>$ (20\%), and  622 $<$A(1-6)$>$ (65\%).\footnote{As outlined in the previous section, when there are multiple ambiguous referents we added an additional identifier (1-...) to indicate whether a sentence contains multiple ambiguous entities as this might have an influence on the amount of different gendered translations.}


\subsection{Multiple Translations}
Sentences that contain ambiguous referents, sometimes -- depending on the target language -- entail multiple equivalent translations in terms of gender. For 157 out of the 695 sentences annotated, this was the case and multiple gender alternative translations were provided in Italian.\footnote{Annotations were provided by linguists and the Italian translations were generated by a native Italian speaker/linguist.}

\section{Analysis and Discussion}

This section provides an analysis and discussion of the specific problems posed by the Italian language and the specific translation choices taken with respect to the gender translations added to the source sentences (see Section \ref{subsec:word-leveltags}). 
First of all, Italian is a pro-drop language and the subject pronoun is often omitted. Therefore in sentences where there are ambiguous subjects (I, you, we, they), like in:
\begin{exmp}
\label{ex:pronoun1}``Why did I $<$A$>$ send her home?''
\end{exmp}
\noindent
there is no need to produce alternative gender translations. However, if there are adjectives referring to ambiguous pronouns, gendered translations are needed, e.g.:
\begin{exmp}
\label{ex:pronoun2} ``A college classmate $<$F$>$ of mine wrote me $<$A$>$ a couple weeks ago and said she thought $<$A$>$ was a little strident.''
\end{exmp}
\noindent
for which we have two Italian sentences, namely ``Una mia ex collega del college mi ha scritto un paio di settimane fa e mi ha detto che mi considerava un po' \textit{eccessivo}.'' for the masculine form and ``Una mia ex collega del college mi ha scritto un paio di settimane fa e mi ha detto che mi considerava un po' \textit{eccessiva}.'' for the feminine form. 
\noindent
The same applies when there are past participle forms in the sentence, since in Italian these forms sometimes require gender agreement with the noun they refer to such as in:

\begin{exmp}
\label{ex:pronoun3} ``I  $<$A$>$ 'd gone to see Pete $<$M$>$, who $<$M$>$ was renowned for his workmanship with steel.''
\end{exmp}

\noindent
for which we produce the alternate gender translation: ``Ero \textit{andato} a vedere Pete, rinomato per la sua abilità nel lavorare l'acciaio.'' and ``Ero \textit{andata} a vedere Pete, rinomato per la sua abilità nel lavorare l'acciaio.''.

Gender translations were needed for bigender Italian nouns as well, such as for instance \textit{insegnante} (teacher) or \textit{paziente} (patient), which have a single invariable form for masculine and feminine and the gender becomes apparent only when there is a coordinated article or adjective , such as in  

\begin{exmp}
\label{ex:bigender} ``Having completed the course, the student $<$M$>$ fed back to us $<$A1$>$ that he now realized that it was he who $<$M$>$ hadn't understood what the client $<$A2$>$ wanted.''
\end{exmp}
\noindent
for which the sentences ``Dopo aver completato il corso, lo studente ci ha riferito che ora ha compreso che era lui a non capire ciò che \textit{il cliente} voleva." and "Dopo aver completato il corso, lo studente ci ha riferito che ora ha compreso che era lui a non capire ciò che \textit{la cliente} voleva..

We also made a conscious decision in terms of the Italian `non-marked' masculine form, also called the inclusive masculine - when the masculine form is used to refer, generically to males and females, such as for instance the use of the masculine form \textit{bambini} (children) to refer to both male and female children. For this particular form, although the use of the inclusive masculine is acceptable when refering to a group of people whose gender is unknown (as proposed in e.g. \citet{robustelli2012linee}), we still opted to provide an alternative translation. For instance the sentence:  
\begin{exmp}
\label{ex:inclusive} ``Man, I $<$A1$>$ come home from work, drawers are open, clothes hanging outside the drawers, the kids $<$A2$>$ are still in their pajamas...''
\end{exmp}
\noindent
is translated as: ``Amico, torno a casa dal lavoro, i cassetti sono aperti, i vestiti tutti fuori, \textit{i bambini} sono ancora in pigiama..." and "Amico, torno a casa dal lavoro, i cassetti sono aperti, i vestiti tutti fuori, \textit{le bambine} sono ancora in pigiama...''

The generic masculine form is also used for the agreement of adjectives/past participles/nouns in agreement with the natural gender of referents that have different genders, e.g.: ``Giovanni e Lucia sono \textit{bravi insegnanti}'' (Giovanni and Lucia are good teachers). In this case, we kept the masculine form as no possible alternatives are currently accepted. Recently, the use of the schwa, ``\textschwa'', was  proposed \cite{gheno2019femminili}, precisely to solve these types of problems related to the use of the inclusive masculine form but also to take into account non-binary people representation needs, nevertheless this solution has not yet been widely adopted and is not accepted as a linguistic norm.

A further problem addressed in providing gender translation is related to the so-called agentive nouns, namely those nouns that are used to classify people that have specific functions, roles, professions. This type of nouns represent the main problem of sexism in the Italian language, and it is currently widely debated, since the tendency is to use male forms also to refer to professions or roles played by women. This is especially true for nouns which refer to particularly prestigious roles, such as \textit{direttore} (director), \textit{presidente} (president), \textit{ministro} (minister), \textit{professore} (professor) and the like, for which feminine nouns exist: \textit{direttrice} (female director), \textit{presidentessa} (female president), \textit{ministra} (female minister), \textit{professoressa} (female professor), etc. These forms are not always used (including by women) as some consider the feminization of a profession a loss of prestige.\footnote{Recently, the Accademia della Crusca, one of the most important research bodies for the Italian Language, discussed this problem with reference to the request by Beatrice Venezi to be presented as ``direttore d’orchestra'' (orchestra director) and not as  ``direttrice d’orchestra'' (female orchestra director) during an important Italian song contest, namely the 71st Festival of Sanremo: \url{https://accademiadellacrusca.it/it/consulenza/direttori-dorchestra-e-maestri-del-coro-anche-se-donne/2917}} For these cases, we opted to provide both masculine and feminine translations:
\begin{exmp}
\label{ex:agentive} ``So I $<$A1$>$ one day decided to pay a visit to the manager $<$A2$>$, `Is this model of offering people $<$A3$>$ all this choice really working?'.''
\end{exmp}
\noindent
for which we provide the following alternate translations: ``E così un giorno decisi di andare a trovare \textit{il direttore}, e \textit{gli} domandai: `Funziona davvero offrire tutta questa scelta alla gente?''' and ``E così un giorno decisi di andare a trovare \textit{la direttrice}, e \textit{le} domandai: `Funziona davvero offrire tutta questa scelta alla gente?''' 

For 23 sentences, more than two alternative translations had to be provided due to multiple ambiguous referents. For instance, for the following sentence:
\begin{exmp}
\label{ex:multiple_trans}
``And when these individuals went and looked at who $<$A1$>$ was the best protein folder $<$A1$>$ in the world, it wasn't an MIT professor $<$A2$>$, it wasn't a CalTech student $<$A3$>$, it was a person $<$F$>$ from England, from Manchester, a woman who $<$F$>$ , during the day, was an executive assistant $<$F$>$ at a rehab clinic and, at night, was the world's best protein folder $<$F$>$ .'' 
\end{exmp}
\noindent
there are 8 alternative translations in terms of gender that can be generated in Italian.

\section{Conclusions and Future Work}
In this paper, we present and describe gENder-IT: an English-Italian annotated parallel challenge set. The English source side is annotated with word-level gender tags, while for the Italian target side the translations --including correct gender alternatives-- are provided. We present a detailed description of the annotations as well as a contrastive analysis of translation specific gender challenges for English--Italian. In future work, we envisage working on: (i) an extension of the corpus to other languages, (ii) the identification of gender ambiguous sentences in English, and (iii) the subsequent generation of multiple gender alternatives where necessary, including paraphrases to adopt more gender-neutral solutions. With our challenge set and analysis, we hope to encourage research on ambiguity detection and the controlled generation of gender diverse alternatives for translations.

\section*{Acknowledgments}
We would like to thank the reviewers for their insightful feedback and comments.

\bibliography{acl2020}
\bibliographystyle{acl_natbib}




\end{document}